\documentclass[conference,letterpapaer]{IEEEtran}

\usepackage{balance}
\usepackage{subfigure}
\usepackage{amssymb}
\usepackage{color}

\usepackage[utf8]{inputenc} 
\usepackage[T1]{fontenc}
\usepackage{url}              
\usepackage{cite}             

\usepackage[cmex10]{amsmath}  
\usepackage{mleftright}       
\mleftright                   

\usepackage{graphicx}         
\usepackage{booktabs}         

\newtheorem{lemma}{Lemma}
\newtheorem{remark}{Remark}
\newtheorem{corollary}{Corollary}
\newtheorem{theorem}{Theorem}
\newtheorem{assumption}{Assumption}

\usepackage{xspace}
\usepackage{bbm}
\usepackage{mathrsfs}

\newcommand{\Ac}{\mathcal{A}}
\newcommand{\Bc}{\mathcal{B}}
\newcommand{\Cc}{\mathcal{C}}

\newcommand{\Dc}{\mathcal{D}}

\newcommand{\Sc}{\mathcal{S}}

\newcommand{\Xc}{\mathcal{X}}
\newcommand{\Yc}{\mathcal{Y}}

\newcommand{\Dv}{{\bf D}}

\def\a{\alpha}

\def\eps{\epsilon}

\DeclareMathOperator\E{\mathbb E}
\let\P\relax
\DeclareMathOperator\P{\mathbb P}

\def\textiid{i.i.d.\@\xspace}
\newcommand\iid{\ifmmode\text{ i.i.d. } \else \textiid \fi}

\newcommand{\Real}{\mathbb{R}}

\newcommand{\ind}{\boldsymbol{1}}

\newcommand{\indep}{\perp \!\!\! \perp}

\begin{document}


\title{Training-Conditional Coverage Bounds for Uniformly Stable Learning Algorithms} 


\author{\IEEEauthorblockN{Mehrdad Pournaderi and Yu Xiang}
\IEEEauthorblockA{\textit{Department of Electrical and Computer Engineering}\\
\textit{University of Utah}\\
Salt Lake City, UT 84112, USA \\
\{m.pournaderi,\,yu.xiang\}@utah.edu}
}

\maketitle

\begin{abstract}
The training-conditional coverage performance of the conformal prediction is known to be empirically sound. Recently, there have been efforts to support this observation with theoretical guarantees. The training-conditional coverage bounds for jackknife+ and full-conformal prediction regions have been established via the notion of $(m,n)$-stability by Liang and Barber~[2023]. Although this notion is weaker than uniform stability, it is not clear how to evaluate it for practical models. In this paper, we study the training-conditional coverage bounds of full-conformal, jackknife+, and CV+ prediction regions from a uniform stability perspective which is known to hold for empirical risk minimization over reproducing kernel Hilbert spaces with convex regularization. We derive coverage bounds for finite-dimensional models by a concentration argument for the (estimated) predictor function, and compare the bounds with existing ones under ridge regression.


\end{abstract}

\section{Introduction and Problem Formulation}

Conformal prediction is a framework for constructing \emph{distribution-free} predictive confidence regions as long as the training and test data are exchangeable~\cite{vovk2005algorithmic} (also see~\cite{shafer2008tutorial,vovk2009line,vovk2012conditional}). Specifically, let $\Dc_{n}\cup (X_{\text{test}}, Y_{\text{test}})$ denote a dataset with exchangeable data points, consisting of a training set of $n$ samples $\Dc_{n}:=\{(X_i,Y_i)\in\Xc\times\Yc:i\in[n]\}$ and one test sample $(X_{\text{test}}, Y_{\text{test}})$, where $[n]:=\{1, 2,..., n\}$. The conformal prediction provides a coverage of $Y_{\text{test}}$ in the sense of 
\begin{equation}
    \P(Y_{\text{test}}\in \hat{C}_{\alpha}(X_{\text{test}}))\geq 1-\alpha,~\label{eq:full}
\end{equation}
where $\hat{C}_{\alpha}:\Xc \to 2^\Yc$ is a data-dependent map.
This type of guarantee is referred to as \emph{marginal} coverage, as it is averaged over all the training and test data. One natural direction to stronger results is to devise \emph{conditional} coverage guarantee
\[
    \P(Y_{\text{test}}\in\hat\Cc_{\a}(X_{\text{test}})|X_{\text{test}})\ge 1-\alpha.
\] 
However, it has been shown in~\cite{vovk2012conditional,foygel2021limits,lei2014distribution} that it is \emph{impossible} to obtain (non-trivial) distribution-free prediction regions $\hat\Cc(x)$ in the finite-sample regime; relaxed versions of this type of guarantee have been extensively studied (see~\cite{jung2022batch,gibbs2023conformal,vovk2003mondrian} and references therein). As an alternative approach, several results~(e.g., \cite{vovk2012conditional,bian2023training}) have been reported on the \emph{training-conditional} guarantee by conditioning on $\Dc_n$, which is also more appealing than the marginal guarantee as can be seen below. Define the following miscoverage rate as a function of the training data,
\[
P_e(\Dc_n):=\P(Y_{\text{test}}\notin\hat\Cc(X_{\text{test}})|\Dc_n).
\] 
Note that the marginal coverage in~\eqref{eq:full} is equivalent to $\E[P_e(\Dc_n)]\le \alpha$. The training-conditional guarantees are of the following form, for some small $\delta$,
\[
\P(P_e(\Dc_n)\geq\alpha)\leq \delta
\]
or its asymptotic variants. Roughly speaking, this guarantee means that the $(1-\alpha)$-level coverage lower bounds hold for a \emph{generic} dataset. 


 In this line of research, samples are assumed to be i.i.d., which is not only exchangeable but also ergodic and admits some nice concentration properties. For the \emph{$K$-fold CV+} with $m$ samples in each fold, the conditional coverage bound
\begin{align}
    \P\left(P_e(\Dc_n)\geq 2\alpha+\sqrt{2\log(K/\delta)/m}\right)\leq \delta \label{cvbian}
\end{align} is established in~\cite{bian2023training}. They have also shown that distribution-free training-conditional guarantees for full-conformal and \emph{jackknife+} methods are impossible without further assumptions; in particular, they conjectured that a certain form of algorithmic stability is needed for full-conformal and jackknife+. Recently, \cite{liang2023algorithmic} proposed (asymptotic) conditional coverage bounds for jackknife+ and full-conformal prediction sets under the assumption that the training algorithm is symmetric. The bound, however, depends on the distribution of the data through the so-called $(m,n)$-stability parameters, where the convergence rate can be slow (see Section~\ref{ridg}). 

This work is motivated by a large class of regression models that can be written as finite-dimensional empirical risk minimization over a reproducing kernel Hilbert space with regularization, i.e., $\hat\mu_{\Dc_n}=g_{\hat\theta_n}$ with
\[
    \hat\theta_n = \underset{\theta\in\Real^p}{\arg\min}\frac{1}{n}\sum_{i\in [n]}\ell(g_\theta(X_i),Y_i)+ \lambda \|g_\theta\|^2,
\]
where $g_\theta,\ \theta\in\Real^p$ is a family of predictor functions parametrized by $\theta$ 
and $\ell$ is some suitable loss function. These models are known to be \emph{uniformly stable}~\cite{bousquet2002stability} in the sense that 
\begin{align}
    \|\hat\mu_{\Dc_n}-\hat\mu_{\Dc'_n}\|_\infty\leq \beta~\label{eq:uni}
\end{align} 
with $\beta=O(1/n)$ for any two datasets $(\Dc_n$, $\Dc'_n)$ that differ in one data sample, which is a stronger notion than the stability assumed in\cite{liang2023algorithmic}. We aim to improve the training-conditional coverage guarantees of these learning models by establishing better rates of convergence for full-conformal and jackknife+. 

\section{Background and Related Work}
\subsection{Full-Conformal and Split-Conformal}
Let $T$ denote a symmetric training algorithm, i.e., the predictor function $\hat\mu:\Xc\to\Yc$ is invariant under permutations of the training data points, and $\hat\mu_{(x,y)}:=T(\Dc_n\cup (x,y))$ is a regression fucntion by running $T$ on $\Dc_n\cup (x,y)$. 
Define the score function $s(x',y';\hat\mu_{(x,y)}):=f(\hat\mu_{(x,y)}(x'),y')$ via some arbitrary (measurable) cost function $f$. For instance, $s(x',y';\hat\mu_{(x,y)})=|y'-\hat\mu_{(x,y)}(x')|$ when $f(y,y')=|y-y'|$. 
Let 
\begin{align*}
   \Sc(x,y;\Dc_n):=\{s(x',y';\hat\mu_{(x,y)}):(x',y')\in \Dc_n\cup (x,y)\} 
\end{align*}
and observe that the elements of $\Sc(X_{\text{test}},Y_{\text{test}};\Dc_n)$
are exchangeable. Therefore,
\begin{align*}
    \P\Big(s(X_{\text{test}},Y_{\text{test}}&;\hat\mu_{(X_{\text{test}},Y_{\text{test}})})\leq \\&\hat{F}^{-1}_{\Sc(X_{\text{test}},Y_{\text{test}};\,\Dc_n)}(1-\alpha)\Big) \geq 1-\alpha,
\end{align*}
where $\hat{F}^{-1}_{\Sc(X_{\text{test}},Y_{\text{test}};\,\Dc_n)}(1-\alpha)$ denotes the \emph{empirical} quantile function with respect to the set of values $\{\Sc(X_{\text{test}},Y_{\text{test}};\,\Dc_n)\}$. Thus, 
\[
    \P(Y_{\text{test}}\in \hat{C}_{\alpha}(X_{\text{test}}))\geq 1-\alpha,
\]
where the following confidence region is referred to as \emph{full-conformal} in the literature
\[
\hat C_\alpha(x) = \{y:s(x,y;\hat\mu_{(x,y)})\leq \hat{F}^{-1}_{\Sc(x,y;\Dc_n)}(1-\alpha)\}.
\] 
It is well-known that this approach can be computationally intensive when $\Yc = \mathbb R$ since to find out whether $y\in \hat C_\alpha(x)$ one needs to train the model with the dataset including $(x,y)$ with $y\in\Real$. One simple way to alleviate this issue is to split the data into training and calibration datasets, namely $\Dc_n = \Dc^{\text{train}}\cup \Dc^{\text{cal}}$. First one finds the regression $\hat\mu:=T(\Dc^{\text{train}})$ and treats $\hat\mu$ as fixed. Let $\tilde\Sc(\Dc_n):=\{s(x,y;\hat\mu):(x,y)\in \Dc_n\}$, and note that the elements of
$\tilde\Sc((X_{\text{test}},Y_{\text{test}})\cup\Dc^{\text{cal}})$ are exchangeable. Hence, we get
\[
    \P\left(s(X_{\text{test}},Y_{\text{test}};\hat\mu)\leq \hat{F}^{-1}_{\tilde\Sc\left((X_{\text{test}},Y_{\text{test}})\cup\Dc^{\text{cal}}\right)}(1-\alpha)\right) \geq 1-\alpha.
\]
Hence, 
\[
    \P(Y_{\text{test}}\in \hat{C}^{\text{split}}_{\alpha}(X_{\text{test}}))\geq 1-\alpha,
\]
for 
\begin{align*}
    \hat C^{\text{split}}_\alpha(x) &=\left\{y:s(x,y;\hat\mu)\leq \hat{F}^{-1}_{\tilde\Sc(\Dc^{\text{cal}})\cup \{\infty\}}(1-\alpha)\right\}\\ &\supseteq\left\{y:s(x,y;\hat\mu)\leq \hat{F}^{-1}_{\tilde\Sc((x,y)\cup\Dc^{\text{cal}})}(1-\alpha)\right\}.
\end{align*}

\subsection{Jackknife+}
Although the split-conformal approach resolves the computational efficiency problem of the full-conformal method, it is somewhat inefficient in using the data and may not be useful in situations where the number of samples is limited. A heuristic alternative has long been known in the literature, namely, jackknife or leave-one-out cross-validation that can provide a compromise between the full conformal and split conformal methods. In particular, 
\[
    \hat C^{\text{J}}_\alpha(x) =\{y:s(x,y;\hat\mu)\leq \hat{F}^{-1}_{\Sc^{\text{cal}}}(1-\alpha)\}
\]
where $\Sc^{\text{cal}}:=\{s(X_i,Y_i;\hat\mu^{-i}): 1\leq i\leq |\Dc^{\text{train}}|\}$ and $\hat\mu^{-i}:=T(\Dc^{\text{train}}\setminus \{(X_i,Y_i)\})$. Despite its effectiveness, no general finite-sample guarantees are known for jackknife. Recently, \cite{barber2021predictive} proposed jackknife+, a modified version of the jackknife for $\Yc=\mathbb R$ and $f(y,y')=|y-y'|$, and established $(1-2\alpha)$ coverage lower bound for it. 
Let $\hat{q}^+_\alpha(A)$ and $\hat{q}^-_\alpha(A)$ denote the $\lceil(1-\alpha)(|A|+1)\rceil$-th 
and $\lfloor \alpha(|A|+1)\rfloor$-th smallest values of the set $A$, respectively, with the convention $\hat{q}^+_\alpha(A)=\infty$ if $\alpha < 1/(n+1)$. Let 
\begin{align*}
    \Sc^{-}(x)&=\{\hat\mu^{-i}(x)-|Y_i-\hat\mu^{-i}(X_i)|:i\in[n]\},\\
    \Sc^{+}(x)&=\{\hat\mu^{-i}(x)+|Y_i-\hat\mu^{-i}(X_i)|:i\in[n]\},
\end{align*}
and the jackknife+ prediction interval is defined as
\[
    \hat C^{\text{J+}}_\alpha(x) = [\hat{q}^-_\alpha(\Sc^{-}(x)),\; \hat{q}^+_\alpha(\Sc^{+}(x))].
\]
In the same paper, an $\eps$-inflated version of the jackknife+ 
\begin{equation}
    \hat C^{\text{J+},\eps}_\alpha(x) = [\hat{q}^-_\alpha(\Sc^{-}(x))-\eps,\; \hat{q}^+_\alpha(\Sc^{+}(x))+\eps]~\label{eq:inflate}
\end{equation} 
is proposed which has $1-\alpha-4\sqrt{\nu}$ coverage lower bound, instead of $1-2\alpha$, if the training procedure satisfies
\[
\max_{i\in [n]}\P(|\hat\mu(X_{\text{test}})-\hat\mu^{-i}(X_{\text{test}})|>\eps)<\nu.
\]
Also, the jackknife+ has been generalized to CV+ for $K$-fold cross-validation, and $(1-2\alpha-\sqrt{2/|\Dc^{\text{train}}|})$ coverage lower bound is established.

\subsection{Asymptotic Training Conditional Coverage~\cite{liang2023algorithmic}}
The bounds established in~\cite{liang2023algorithmic} depend on the distribution of the data through the $(m,n)$-stability parameters,
\begin{align}
    \psi_{m,n}^{\text{out}}&=\E_{\Dc_{n+m}}|\hat\mu_{\Dc_n}(X_{\text{test}})-\hat\mu_{\Dc_{n+m}}(X_{\text{test}})|,\label{eq:out}\\  \psi_{m,n}^{\text{in}}&=\E_{\Dc_{n+m}}|\hat\mu_{\Dc_n}(X_1)-\hat\mu_{\Dc_{n+m}}(X_1)|.
\end{align}
where $\hat\mu_{\Dc_n}=T(\Dc_n)$ and $X_{\text{test}}\indep \Dc_{n+m}$ with $\Dc_{n+m} =\{(X_1,Y_1), ..., (X_{n+m}, Y_{n+m})\}$. Tight bounds for this parameter are not known yet. Therefore, the current convergence rates appear to be slow in sample size --- see in Section~\ref{ridg} for details. 
Furthermore, in this analysis, a $\gamma$-inflated version (as in~\eqref{eq:inflate}) of the method is considered, hence, one needs to deal with terms of the form $(\psi_{m,n}/\gamma)^{1/3}$ in the bound which can make the rates even slower if one let $\gamma \to 0$.  We aim to improve the training-conditional coverage guarantees of these learning models in the following ways: (1)~establishing $n^{-1/2}$ rates with explicit dependence on the dimension of the problem and (2)~removing the interval inflation.

\section{Conditional Coverage Guarantees}
Let $\mu_\beta\in L^{\infty}(\Xc)$ denote a predictor function parameterized by $\beta\in\Real^p$. By a slight abuse of notation, let the map $T:\cup_{n\geq 1}(\Xc\times \Yc)^n\to \Real^p$ denote a training algorithm for estimating $\beta$, hence, $\hat\beta_n=T(\Dv_n)$ where $\Dv_n:=((X_1,Y_1),\hdots,(X_n,Y_n))\in (\Xc\times \Yc)^n$ denotes the $\iid$ training tuple of data points. In this case, we have $\hat\mu_{\Dv_n}=\mu_{\hat\beta_n}$.
\smallskip
\smallskip
\begin{assumption}[Uniform stability] For all $i\in [n]$, we have
    \[
        \underset{z_1,\hdots,z_n}{\sup}\|\mu_{T(z_1,\hdots,z_{i-1},z_{i+1}\hdots,z_n)}-\mu_{T(z_1,\hdots,z_i,\hdots, z_n)}\|_\infty\leq \frac{c_n}{2}.
    \]
     \label{model_stab}
\end{assumption}

In the case of the ridge regression~\cite{hoerl1970ridge} with $\Yc=[-B,B]$ and $\Xc=\{x:\|x\|_2\leq b\}$, this assumption holds with $c_n={16\, b^2 B^2}/{(\lambda \, n)}$ where $\lambda$ denotes the regularization parameter~\cite{bousquet2002stability}.
\smallskip
\begin{assumption} The model is bi-Lipschitz (Lipeomorphism) in parameters,
 \[
    \kappa_1 \big\|\beta-\beta'\big\|_\infty\leq \big\|\mu_{\beta}-\mu_{\beta'}\big\|_\infty\leq \kappa_2 \big\|\beta-\beta'\big\|_\infty, 
 \]
 with $\kappa_1 > 0$ and $\kappa_2 < \infty$.
 \label{bilip}
\end{assumption}

\smallskip
\begin{remark}
It is worth noting that if the parameter space $\Theta$ is compact, $\Phi:U\to L^{\infty}(\Xc)$ given by $\beta \mapsto \mu_\beta$ is continuously differentiable for some open $U\supseteq \Theta$, then $\kappa_2 < \infty$. Moreover, the inverse function theorem (for Banach spaces), gives the sufficient condition under which the inverse is continuously differentiable over $\Phi(U)$ and hence $\kappa_1 > 0$.    
\end{remark}

\smallskip
In the case of linear regression with $\Xc=\{x:\|x\|_2\leq b\}$, one can verify that Assumption~\ref{bilip} holds with $\kappa_1 = b$ and $\kappa_2=\sqrt{p} b$. 

Let $\overline{\beta}_n=\E\hat\beta_n$, $\hat\beta_{-i}=T(Z_1,\hdots,Z_{i-1},Z_{i+1},\hdots,Z_n)$ where $Z_i=(X_i,Y_i)$, and $\overline{\beta}_{-i}=\E\hat\beta_{-i}$.
Define 
\begin{align*}
    F_{n-1}(t)&:=\P\left(\left|Y_1-\mu_{\overline{\beta}_{-1}}(X_1)\right|\leq t\right),\\
      \hat{F}_{n-1}(t)&:=\frac{1}{n}\sum_{i=1}^n \ind\left\{\left|Y_i-\mu_{\overline{\beta}_{-1}}(X_i)\right|\leq t\right\}.
\end{align*}

\smallskip
\begin{assumption}[Bounded density]
    $F_n'<L_n$ .\label{bdens}
\end{assumption}

\smallskip
\begin{theorem}[Jackknife+]
Under Assumptions~\ref{model_stab}---\ref{bdens}, for all $\eps,\delta > 0$, it holds that
\begin{align*}
    \P\Bigg( P_e^{\,\text{J+}} & (\Dv_n)  > \alpha+\sqrt{\frac{\log(2/\delta)}{2n}}+\\ & 2\, L_{n-1}\, \kappa_2 \,c_{n-1}\bigg(\frac{1}{\kappa_1}+\sqrt{\frac{n}{2 \kappa_1^2}\log\frac{2p}{\eps}}\bigg)\Bigg)\leq \eps+\delta.
\end{align*}
\end{theorem}
\smallskip

Using the same arguments as in the proof of this theorem, one can get a coverage bound for the CV+ as well. Unlike~\eqref{cvbian} which is meaningful only if the number of samples in each fold $m$ is large, the bound we present in the following corollary is suitable for cases where $m/n\to 0$.  

\smallskip
\begin{corollary}[CV+]
 Under Assumptions~\ref{model_stab}---\ref{bdens}, for all $\eps,\delta > 0$, it holds that
\begin{align*}
    \P\Bigg(& P_e^{\,\text{CV+}}  (\Dv_n)  > \alpha+\sqrt{\frac{\log(2/\delta)}{2n}}+\\ & 2\, m\, L_{n-m}\, \kappa_2\, c_{n-m}\bigg(\frac{1}{\kappa_1}+\sqrt{\frac{n}{2 \kappa_1^2}\log\frac{2p}{\eps}}\bigg)\Bigg)\leq \eps+\delta.
\end{align*}
\end{corollary}

\smallskip
The following theorem concerns the training-conditional guarantees for the full-conformal prediction regions.

\smallskip
\begin{theorem}[Full-conformal]
Under Assumptions~\ref{model_stab}---\ref{bdens}, for all $\eps,\delta > 0$, it holds that
\begin{align*}
    \P\Bigg(P_e(\Dv_n)  > &\, \alpha+\sqrt{\frac{\log(2/\delta)}{2n}}+\\ & L_n\left(c_{n+1}+\sqrt{2n \log\frac{2p}{\eps}}\,\frac{ \kappa_2 \, c_n}{\kappa_1}\right)\Bigg)\leq \eps+\delta.
\end{align*}
\end{theorem}

\section{Coverage Bounds for Ridge Regression} \label{ridg}
In this section, we wish to evaluate the bounds for the ridge regression with $\Xc=\{x:\|x\|\leq b\}$ and $\Yc=[-B,B]$. As stated in the previous section, this regression model satisfies $c_n={16\, b^2 B^2}/{(\lambda\, n)}$, $\kappa_1 = b$ and $\kappa_2=\sqrt{p}\, b$. Hence, we get the following bound for both full-conformal and jackknife+ methods,
\begin{align*}
    \P\Bigg( P_e(\Dv_n)  > \alpha+O\bigg(n^{-1/2}\bigg(\sqrt{{\log(\frac{1}{\delta})}}+ & \sqrt{p\log(\frac{2p}{\eps})}\bigg)\bigg)\Bigg)\\ &\leq \eps+\delta.
\end{align*}
On the other hand, the following bound is proposed for the $\gamma$-inflated jackknife in \cite{liang2023algorithmic},
\begin{align}
    \P\Bigg( P_e^{\text{J+},\gamma} (\Dv_n)  > \alpha+ 3\sqrt{\frac{\log(1/\delta)}{\min(m,n)}}+ & 2\sqrt[3]{\frac{\psi_{m,{n-1}}^{\text{out}}}{\gamma}}\Bigg) \label{dimfree}\\ &\leq 3\delta+\sqrt[3]{\frac{\psi_{m,{n-1}}^{\text{out}}}{\gamma}}.\nonumber
\end{align}
for all $m\geq 1$.
We get $\psi_{m,n}^{\text{out}} = O(m c_n)$ since $\psi_{1,n}^{\text{out}}\leq c_{n+1}/2$ by definition~\eqref{eq:out} and Assumption~\ref{model_stab}, and $\psi_{m,n}^{\text{out}} \leq \sum_{k=n}^{n+m-1}\psi_{1,k}^{\text{out}}$ holds according to in \cite[Lemma~5.2]{liang2023algorithmic} . Substituting for $\psi_{m,{n-1}}^{\text{out}}$ in bound~\eqref{dimfree}, we obtain
 \begin{align*}
    \P\Bigg( P_e^{\text{J+},\gamma} (\Dv_n)  > \alpha+ O \bigg( & \sqrt{\frac{\log(1/\delta)}{\min(m,n)}}+  \sqrt[3]{\frac{m\, c_{n-1}}{\gamma}}\bigg)\Bigg)\\ &\leq 3\delta+O\left(\sqrt[3]{\frac{m c_{n-1}}{\gamma}}\right).\nonumber
\end{align*}
Letting $m^{-1/2} = (m/n)^{1/3}$ to balance the two terms $\sqrt{\frac{\log(1/\delta)}{\min(m,n)}}$ and $\sqrt[3]{{m c_{n-1}/\gamma}}$, we get $m=n^{2/5}$. By plugging $m= n^{2/5}$ in, we get
 \begin{align}
    \P\Bigg( P_e^{\text{J+},\gamma} (\Dv_n)  > \alpha+ O \bigg( & n^{-1/5}\left(\sqrt{{\log(1/\delta)}}+  {\gamma}^{-1/3}\right)\bigg)\Bigg)\nonumber\\ &\leq 3\delta+O\left(n^{-1/5}{\gamma}^{-1/3}\right). \label{ridgebnd}
\end{align}
This bound, although dimension-free, is very slow in the sample size. In~\cite{liang2023algorithmic}, the same bound as~\eqref{dimfree} is established for $\gamma$-inflated full-conformal method except with $\psi_{m-1,n+1}^{\text{in}}$ instead of $\psi_{m,{n-1}}^{\text{out}}$. Hence, the same bound as~\eqref{ridgebnd} can be obtained for the $\gamma$-inflated full-conformal method via $\psi_{m,n}^{\text{in}}=O(m c_n)$.
\section{conclusion}
The $(m,n)$-stability is a new measure of the stability of a regression model. It was recently introduced in~\cite{liang2023algorithmic} and used to compute training-conditional coverage bounds for full-conformal and jackknife+ prediction intervals. Unlike uniform stability which is a distribution-free property of a training process, $(m,n)$-stability depends on both the training algorithm and the distributions of the data. Although weaker than uniform stability, the parameter is not well-understood in a practical sense yet. In this work, we have studied the training-conditional coverage bounds of full-conformal, jackknife+, and CV+ prediction regions from a uniform stability perspective which is well understood for convexly regularized empirical risk minimization over reproducing kernel Hilbert spaces. We have derived new bounds via a concentration argument for the (estimated) predictor function. In the case of ridge regression, we have used the uniform stability parameter to derive a bound for the $(m,n)$-stability and compare the resulting bounds from~\cite{liang2023algorithmic} to the bounds established in this paper. We have observed that our rates are faster in sample size but dependent to the dimension of the problem.
\balance
\bibliographystyle{IEEEtran}
\bibliography{ref.bib}

\newpage
\appendices
\onecolumn
\section{Proof for jackknife+}
 
\begin{lemma}
    If Assumption~\ref{model_stab} and~\ref{bilip} hold, then
    \[
        \P\Big(\left\|\hat\beta_n -\E\hat\beta_n\right\|_\infty\geq \eps\Big)\leq 2p \exp\left({-\frac{2 \kappa_1^2 \eps^2}{nc_n^2}}\right).
    \]
    \label{conc}
\end{lemma}
\begin{IEEEproof}
     Assumption~\ref{model_stab} and~\ref{bilip} imply that
\begin{align*}
        \underset{z_1,\hdots,z_n,z_i'}{\sup}\|T(z_1,\hdots,z_i\hdots,z_n)-T(z_1,\hdots, z_i',\hdots,& z_n)\|_\infty\leq \frac{c_n}{\kappa_1}.   
\end{align*}
By McDiarmid's inequality~\cite{mcdiarmid1989method} we get
\begin{align}
    \P\Big(\|\hat\beta_n &-\E\hat\beta_n\|_\infty\geq \eps\Big)=\P\Big(\big\|T(Z_1,\hdots,Z_n)-\E\,T(Z_1,\hdots,Z_n)\big\|_\infty\geq \eps\Big)
    \leq 2p \exp\left({-\frac{ 2\kappa_1^2\eps^2}{nc_n^2}}\right)
\end{align}
for independent $Z_i$ and all $\eps>0$. 
\end{IEEEproof}
\smallskip
\smallskip
\begin{lemma} Under Assumptions~\ref{model_stab} and \ref{bilip} we have
\begin{align*}
    \P\Big(\max_i \left\|\mu_{\hat\beta_{-i}}-\mu_{\overline{\beta}_{-1}}\right\|_\infty \geq\eps\Big)\leq  2p\exp\left(-\frac{2 \kappa_1^2}{n}\left(\frac{\eps}{\kappa_2 c_{n-1}}-\frac{1}{\kappa_1}\right)^2\right).
\end{align*}\label{model_conc}
\end{lemma}
\begin{IEEEproof}
From Assumption~\ref{model_stab} and \ref{bilip}, it follows that 
\begin{equation}
    \max_{i,j}\|\hat\beta_{-i}-\hat\beta_{-j}\|_\infty\leq \frac{c_{n-1}}{\kappa_1}.\label{cluster}
\end{equation}

Also, according to~\eqref{conc}, we have $\|\hat\beta_{-1}-\overline{\beta}_{-1}\|_\infty<\eps$ with probability at least $1-2p\exp({-{2\kappa_1^2 \eps^2}/(nc_{n-1}^2}))$. 
We note that,
\begin{align*}
    \P\Big(\max_i\Big\|\mu_{\hat\beta_{-i}}  -\mu_{\overline{\beta}_{-1}}\Big\|_\infty \geq\eps\Big) & \overset{(*)}{\leq}\P\left(\kappa_2\max_i\left\|\hat\beta_{-i}-\overline{\beta}_{-1}\right\|_\infty\geq\eps\right)\\
    &\overset{(**)}{\leq} \P\left(\kappa_2\left(\frac{c_{n-1}}{\kappa_1}+\left\|\hat\beta_{-1}-\overline{\beta}_{-1}\right\|_\infty\right)\geq\eps\right)\\
    &\leq 2p\exp\left(-\frac{2 \kappa_1^2}{n}\left(\frac{\eps}{\kappa_2 c_{n-1}}-\frac{1}{\kappa_1}\right)^2\right).
\end{align*}
where (*) and (**) hold according to Assumption~\ref{bilip} and \eqref{cluster}, respectively.
\end{IEEEproof}
\smallskip


Let $\hat\Cc_\alpha(X_{n+1})$ denote the Jackknife+ $\alpha$-level interval for test data-point $X_{n+1}$ and define $P_e(\Dv_n):=\P(Y_{n+1}\notin\hat\Cc(X_{n+1})|\Dv_n)$.
\smallskip
\begin{IEEEproof}
We note,
\begin{align*}
    \hat\Cc_\alpha(X_{n+1})&\supseteq \left\{y\in \Real:\frac{1}{n}\sum_{i=1}^n \ind\left\{\left|Y_i-\mu_{\hat\beta_{-i}}(X_i)\right|\geq \left|y-\mu_{\hat\beta_{-i}}(X_{n+1})\right|\right\}>\alpha\right\}\\
    &\supseteq\Bigg\{y\in \Real:\frac{1}{n}\sum_{i=1}^n \ind\bigg\{\left|Y_i-\mu_{\overline{\beta}_{-1}}(X_i)\right|-\left|\mu_{\hat\beta_{-i}}(X_i)-\mu_{\overline{\beta}_{-1}}(X_i)\right|\geq\\
    &\hspace{120pt}\left|y-\mu_{\overline{\beta}_{-1}}(X_{n+1})\right|+\left|\mu_{\hat\beta_{-i}}(X_{n+1})-\mu_{\overline{\beta}_{-1}}(X_{n+1})\right|\bigg\}>\alpha\Bigg\},
\end{align*}
where the first relation holds according to \cite{bian2023training}.
Assuming $\max_i\|\mu_{\hat\beta_{-i}}-\mu_{\overline{\beta}_{-1}}\|_\infty<\eps$, we obtain
\begin{align*}
    \hat\Cc_\alpha(X_{n+1})
    &\supseteq\Bigg\{y\in \Real:\frac{1}{n}\sum_{i=1}^n \ind\bigg\{\left|Y_i-\mu_{\overline{\beta}_{-1}}(X_i)\right|\geq
   \left|y-\mu_{\overline{\beta}_{-1}}(X_{n+1})\right|+2\eps\bigg\}>\alpha\Bigg\}
   \\&\supseteq\Bigg\{y\in \Real:1-\hat{F}_{n-1}\left(
   \left|y-\mu_{\overline{\beta}_{-1}}(X_{n+1})\right|+2\eps\right)>\alpha\Bigg\}.
\end{align*}
Assuming $\left\|\hat{F}_{n-1}-F_{n-1}\right\|_\infty<\delta$, we obtain
\begin{align*}
    \hat\Cc_\alpha(X_{n+1})
   &\supseteq\Bigg\{y\in \Real:1-F_{n-1}\left(
   \left|y-\mu_{\overline{\beta}_{-1}}(X_{n+1})\right|+2\eps\right)>\alpha+\delta\Bigg\}\\
   &\supseteq\Bigg\{y\in \Real:1-F_{n-1}\left(
   \left|y-\mu_{\overline{\beta}_{-1}}(X_{n+1})\right|\right)>\alpha+\delta+2\eps L\Bigg\}
\end{align*}
Therefore,
\begin{align*}
    P_e(\Dv_n)=\P(Y_{n+1}\notin\hat\Cc(X_{n+1})|\Dv_n)&\leq \P\left(1-F_{n-1}\left(
   \left|Y_{n+1}-\mu_{\overline{\beta}_{-1}}(X_{n+1})\right|\right)\leq\alpha+\delta+2\eps L\right)\\
   & = \alpha+\delta+2\eps L
\end{align*}
for $\Dv_n\in\Ac\cap\Bc$ where $\Ac:=\left\{D:\max_i\|\mu_{\hat\beta_{-i}}-\mu_{\overline{\beta}_{-1}}\|_\infty<\eps\right\}$ and $\Bc:=\left\{D:\left\|\hat{F}_{n-1}-F_{n-1}\right\|_\infty<\delta\right\}$. From Lemma~\ref{model_conc}, we know $\P(\Dv_n\notin\Ac)\leq 2p\exp\left(-\frac{2 \kappa_1^2}{n}\left(\frac{\eps}{\kappa_2 c_{n-1}}-\frac{1}{\kappa_1}\right)^2\right)$. Also, according to Dvoretzky–Kiefer–Wolfowitz inequality~\cite{dvoretzky1956asymptotic}, we have $\P(\Dv_n\notin\Bc)\leq 2e^{-2n\delta^2}$. Thus,
\begin{align*}
    \P(P_e(\Dv_n)>\alpha+\delta+\eps)\leq \P\left((\Ac\cap\Bc)^c\right)\leq 2e^{-2n\delta^2}+ 2p\exp\left(-\frac{2 \kappa_1^2}{n}\left(\frac{\eps}{2 L_{n-1} \kappa_2 c_{n-1}}-\frac{1}{\kappa_1}\right)^2\right),
\end{align*}
or equivalently,
\begin{align*}
    \P\left(P_e(\Dv_n)>\alpha+\sqrt{\frac{\log(2/\delta)}{2n}}+2\, L_{n-1}\, \kappa_2\, c_{n-1}\left(\frac{1}{\kappa_1}+\sqrt{\frac{n}{2 \kappa_1^2}\log\frac{2p}{\eps}}\right)\right)\leq \eps+\delta.
\end{align*}
\end{IEEEproof}
\section{Proof for Full-Conformal}

\begin{lemma} Under Assumptions~\ref{model_stab} and \ref{bilip}, we have
\begin{align*}
    \P\left(\left\|\mu_{\hat\beta_{n}}-\mu_{\overline{\beta}_{n}}\right\|_\infty \geq\eps\right)\leq 2p \exp\left({-\frac{2 \kappa_1^2 \eps^2}{n \kappa_2^2 c_n^2}}\right).
\end{align*}
\label{model_conc2}
\end{lemma}
\begin{IEEEproof}
 According to Lemma~\ref{conc}, we have $\|\hat\beta_{n}-\overline{\beta}_{n}\|_\infty<\eps$ with probability at least $1-2p \exp\left({-\frac{2 \kappa_1^2 \eps^2}{nc_n^2}}\right)$. 
It follows from Assumption~\ref{bilip} that,
\begin{align*}
    \P\left(\left\|\mu_{\hat\beta_{n}}-\mu_{\overline{\beta}_{n}}\right\|_\infty \geq\eps\right)&\leq \P\left(\kappa_2 \left\|\hat\beta_{n}-\overline{\beta}_{n}\right\|_\infty\geq\eps\right)
    \leq 2p \exp\left({-\frac{2 \kappa_1^2 \eps^2}{n \kappa_2^2 c_n^2}}\right).
\end{align*}
\end{IEEEproof}
Let $\hat\Cc_\alpha(X_{n+1})$ denote the full-conformal $\alpha$-level interval for test data-point $X_{n+1}$ and define $P_e(\Dv_n):=\P(Y_{n+1}\notin\hat\Cc(X_{n+1})|\Dv_n)$. Define $\hat\beta_{X_{n+1},y}:=T((X_1,Y_1),\hdots,(X_n,Y_n),(X_{n+1},y))$.
\smallskip

\begin{IEEEproof}
We note,
\begin{align*}
    \hat\Cc_\alpha(X_{n+1})&\supseteq \left\{y\in \Real:\frac{1}{n}\sum_{i=1}^n \ind\left\{\left|Y_i-\mu_{\hat\beta_{X_{n+1},y}}(X_i)\right|\geq \left|y-\mu_{\hat\beta_{X_{n+1},y}}(X_{n+1})\right|\right\}>\alpha\right\}\\
    &\supseteq\Bigg\{y\in \Real:\frac{1}{n}\sum_{i=1}^n \ind\bigg\{\left|Y_i-\mu_{\hat{\beta}_{n}}(X_i)\right|-\left|\mu_{\hat{\beta}_{n}}(X_i)-\mu_{\hat\beta_{X_{n+1},y}}(X_i)\right|\geq\\
    &\hspace{120pt}\left|y-\mu_{\hat{\beta}_{n}}(X_{n+1})\right|+\left|\mu_{\hat{\beta}_{n}}(X_{n+1})-\mu_{\hat\beta_{X_{n+1},y}}(X_{n+1})\right|\bigg\}>\alpha\Bigg\}\\
    &\supseteq\Bigg\{y\in \Real:\frac{1}{n}\sum_{i=1}^n \ind\bigg\{\left|Y_i-\mu_{\hat{\beta}_{n}}(X_i)\right|\geq\left|y-\mu_{\hat{\beta}_{n}}(X_{n+1})\right|+ c_{n+1}\bigg\}>\alpha\Bigg\},
\end{align*}
where the first and last relations hold according to the definition of $\hat\Cc_\alpha(X_{n+1})$ and Assumption~\ref{model_stab}.
Assuming $\|\mu_{\hat\beta_{n}}-\mu_{\overline{\beta}_{n}}\|_\infty<\eps$, we obtain
\begin{align*}
    \hat\Cc_\alpha(X_{n+1})
    &\supseteq\Bigg\{y\in \Real:\frac{1}{n}\sum_{i=1}^n \ind\bigg\{\left|Y_i-\mu_{\overline{\beta}_{n}}(X_i)\right|\geq
   \left|y-\mu_{\overline{\beta}_{n}}(X_{n+1})\right|+ c_{n+1}+2\eps\bigg\}>\alpha\Bigg\}
   \\&\supseteq\Bigg\{y\in \Real:1-\hat{F}_n\left(
   \left|y-\mu_{\overline{\beta}_{n}}(X_{n+1})\right|+ c_{n+1}+2\eps\right)>\alpha\Bigg\}.
\end{align*}
Assuming $\left\|\hat{F}_n-F_n\right\|_\infty<\delta$, we obtain
\begin{align*}
    \hat\Cc_\alpha(X_{n+1})
   &\supseteq\Bigg\{y\in \Real:1-F_n\left(
   \left|y-\mu_{\overline{\beta}_{n}}(X_{n+1})\right|+c_{n+1}+2\eps\right)>\alpha+\delta\Bigg\}\\
   &\supseteq\Bigg\{y\in \Real:1-F_n\left(
   \left|y-\mu_{\overline{\beta}_{n}}(X_{n+1})\right|\right)>\alpha+\delta+(2\eps+ c_{n+1}) L_n\Bigg\}.
\end{align*}
Therefore,
\begin{align*}
    P_e(\Dv_n)&=\P(Y_{n+1}\notin\hat\Cc(X_{n+1})|\Dv_n)\\
    &\leq \P\left(1-F_n\left(
   \left|Y_{n+1}-\mu_{\overline{\beta}_{n}}(X_{n+1})\right|\right)\leq\alpha+\delta+(2\eps+ c_{n+1}) L_n \right)\\
   & = \alpha+\delta+(2\eps+ c_{n+1}) L_n
\end{align*}
for $\Dv_n\in\Ac\cap\Bc$ where $\Ac:=\left\{D:\|\mu_{\hat\beta_{n}}-\mu_{\overline{\beta}_{n}}\|_\infty<\eps\right\}$ and $\Bc:=\left\{D:\left\|\hat{F}_n-F\right\|_\infty<\delta\right\}$. From Lemma~\ref{model_conc}, we know $\P(\Dv_n\notin\Ac)\leq 2p \exp\left({-\frac{2 \kappa_1^2 \eps^2}{n \kappa_2^2 c_n^2}}\right)$. Also, according to Dvoretzky–Kiefer–Wolfowitz inequality, we have $\P(\Dv_n\notin\Bc)\leq 2e^{-2n\delta^2}$. Thus,
\begin{align*}
    \P(P_e(\Dv_n)>\alpha+\delta+\eps)\leq \P\left((\Ac\cap\Bc)^c\right)\leq 2e^{-2n\delta^2}+ 2p\exp\left(-\left(\frac{\kappa_1(\eps/L_n-c_{n+1})}{\sqrt{2n } \kappa_2 c_{n}}\right)^2\right),
\end{align*}
or equivalently,
\begin{align*}
    \P\left(P_e(\Dv_n)>\alpha+\sqrt{\frac{\log(2/\delta)}{2n}}+L_n\left(c_{n+1}+\sqrt{2n \log\frac{2p}{\eps}}\,\frac{ \kappa_2 \, c_n}{\kappa_1}\right)\right)\leq \eps+\delta.
\end{align*}
\end{IEEEproof}

\balance

\end{document}